\documentclass[11pt]{article}

\usepackage[preprint]{acl}

\usepackage{times}
\usepackage{latexsym}

\usepackage[T1]{fontenc}

\usepackage[utf8]{inputenc}

\usepackage{microtype}

\usepackage{inconsolata}

\usepackage{graphicx}

\usepackage{booktabs}
\usepackage{multirow}
\usepackage{makecell}

\usepackage{amsmath}
\usepackage{amssymb}

\usepackage{tikz}
\usetikzlibrary{shapes, arrows.meta, positioning, shadows, fit, backgrounds, calc}
\usetikzlibrary{shapes.geometric}

\usepackage{subcaption}

\usepackage{listingsutf8}
\usepackage{xcolor}
\definecolor{effBlue}{RGB}{225, 240, 255}
\definecolor{effBorder}{RGB}{100, 149, 237}
\definecolor{qcRed}{RGB}{255, 230, 230}
\definecolor{qcBorder}{RGB}{205, 92, 92}
\definecolor{goldenGreen}{RGB}{85, 107, 47}
\definecolor{conflictRed}{RGB}{178, 34, 34}

\title{Double Triangle Annotation: A Scalable Human-in-the-Loop\\
       Framework for High-Precision Historical Document Annotation}

\author{
  \textbf{Ren Yi} \\
  École Polytechnique Fédérale de Lausanne \\
  \texttt{bonjour@renyi.ch}
}

\begin{document}
\maketitle

% -----------------------------------------------------------------------
\begin{abstract}
Evaluating structured-information extraction from historical documents at scale requires high-precision ground-truth annotations, yet traditional manual labeling is expensive and fully automated pipelines built on large language models are prone to hallucination.
We propose \textbf{Double Triangle Annotation}, a two-layer human-in-the-loop framework that leverages cross-model consensus to automate the majority of annotation work while ensuring high-precision outputs.
In the first layer, two architecturally independent Multimodal Large Language Models annotate each document in parallel; when they agree, the label is auto-accepted, and disagreements are routed to a human jury.
A second layer cross-checks two such systems against each other, escalating residual conflicts to a domain expert.
The framework rests on a single assumption---error independence between models---requires no distributional priors or task-specific calibration, and becomes more autonomous as model capability improves.
On the \emph{Guides Rosenwald}, a corpus of French medical directories spanning 1887--1906, the framework achieves a final Word Error Rate of 0.003.
Applied at scale, model consensus auto-accepts over 85\% of 13{,}595~fields.
We release the resulting benchmark---the first structured-extraction ground truth for the Rosenwald Guides---to support future work on historical document processing.
\end{abstract}

% -----------------------------------------------------------------------
\section{Introduction}
\label{sec:intro}

Historical archives---medical directories, census records, notarial registers---hold structured information that is critical for research in the digital humanities.
Extracting this information at scale requires converting degraded page images into machine-readable records, a task that demands high-precision ground-truth annotations for both training and evaluation.
Yet producing such annotations remains a bottleneck: traditional independent manual annotation is expensive and difficult to scale.
Recent Multimodal Large Language Models (MLLMs) have shown increasingly strong performance on document understanding tasks, but the reliability of any single model remains insufficient for gold-standard annotation due to hallucination.
The fundamental challenge is therefore not whether models can assist annotation, but how to \emph{verify} their outputs efficiently.

We propose the \textbf{Double Triangle Annotation} framework, a two-layer human-in-the-loop paradigm that addresses this challenge through \emph{consensus-based filtering}.
In the first layer, two architecturally independent models annotate each document in parallel; when they agree, the label is auto-accepted, \textbf{based on our assumption that independent strong models are unlikely to produce the same error on the same field}.
Disagreements are routed to a human jury, shifting the annotator's role from producer to verifier.
A second layer cross-checks two such systems against each other, \textbf{reapplying the same assumption}, and escalating residual conflicts to a domain expert.
The design minimizes human effort while inducing a cascaded error reduction that drives the final error rate toward zero.

We evaluate the framework on the \emph{Guides Rosenwald}, a corpus of French medical directories spanning 1887--1906.
In a controlled evaluation on 380~fields, the first-layer juries correct 18.4\% and 26.3\% of fields respectively, and the final reviewer corrects a further 4.2\%, yielding a final Word Error Rate of 0.003.
Applied at scale to 105~sampled pages, model consensus auto-accepts over 85\% of the 13{,}595~fields in the final benchmark.

Our contributions are as follows:
\begin{enumerate}
    \item We introduce the \textbf{Double Triangle Annotation} framework (\S\ref{sec:framework}), a two-layer paradigm that leverages cross-model consensus to automate the majority of annotation work while reserving human effort strictly for ambiguous cases. On our evaluation data, model consensus auto-accepts over 73\% of fields while achieving a final WER of 0.003; applied at scale, the auto-accept rate rises to over 85\%.

    \item The framework rests on a single, intuitive assumption---that independent strong models are unlikely to produce the same error on the same field---requiring no task-specific calibration or prior assumption on annotation distribution. This \textbf{minimal-assumption design} makes it directly applicable to any structured extraction task where two or more models can be run in parallel.

    \item The architecture is \textbf{future-proof}: as model capability improves, the framework will benefit from the resulting increase in consensus, further reducing human effort and producing even higher-precision outputs.
    \item We apply the framework to construct the first \textbf{structured-extraction benchmark} for the Rosenwald Guides (\S\ref{sec:benchmark}), comprising 60~document columns and 13{,}595~annotated fields, and release it to support future work on historical document processing.
\end{enumerate}

\section{Related Work}
\label{sec:related}

\subsection{Historical Document Annotation}

Existing approaches on constructing annotated benchmarks for historical documents converge on a common two-phase paradigm: apply automated processing first, then route the output to human reviewers for correction.
In the simplest variant, projects such as DIVA-HisDB \citep{simistira2016diva}, SCUT-COUCH \citep{yan2010scut}, and the Japanese historical document dataset of \citet{shen2020japanese} employ dedicated pipelines for binarisation, line segmentation, character segmentation, or layout analysis.
Their outputs are then corrected in a single pass by human annotators.

More sophisticated approaches invest in multi-round quality control.
M5HisDoc \citep{NEURIPS2023_m5hisdoc} applies three rounds of strict manual checking after machine-assisted extraction.
CASIA-AHCDB \citep{xu2019casia} adds a classifier-based confidence filter that selectively routes low-confidence items to cross-validation---a form of selective review that reduces effort relative to exhaustive checking, though it relies on a single system's internal confidence rather than agreement between independent systems.
At larger scales, ENP \citep{clausner2015enp} and IMPACT \citep{papadopoulos2013impact} outsource annotation to commercial service providers and enforce structured verification combining automated guideline checks with manual inspection.

Crucially, while methods such as CASIA-AHCDB reduce review volume through single-system confidence filtering, none of these approaches leverage \emph{agreement between independent systems} as a quality signal.
% They also target layout analysis, character segmentation, or text recognition rather than \emph{structured field extraction} from semi-structured documents, which is the focus of the present work.

\subsection{Human--LLM Collaborative Annotation}
\label{subsec:human-llm}

LLMs can substantially reduce annotation cost.
\citet{DBLP:conf/acl/DingQLCLJB23} show that GPT-3 produces usable annotations across a range of tasks at a fraction of human cost, though quality deteriorates on complex tasks involving subtle distinctions or domain-specific knowledge.
The dominant hybrid strategy is \emph{confidence-based selective review}: the LLM labels all items and a confidence score determines which ones are routed to human annotators.
\citet{DBLP:conf/emnlp/WangLXZZ21} report 50--96\% cost reductions with this approach by having humans re-label only the lowest-confidence items; \citet{DBLP:conf/eacl/KimMCRZ24} build a similar pipeline in MEGAnno+.
\citet{DBLP:conf/chi/Wang0RMM24} extend the pattern by deploying two LLMs---one for labeling and one for generating confidence scores---though the quality signal remains single-system confidence rather than inter-model agreement.

However, these workflows introduce systematic risks.
\citet{DBLP:conf/acl/SchroederRK25} find that presenting annotators with LLM-generated labels increases their self-reported confidence but does not accelerate annotation, and introduces \emph{anchoring bias} whereby annotators defer to the model's suggestion.
\citet{DBLP:journals/corr/abs-2503-06778} reach a complementary conclusion: LLMs are effective annotation \emph{assistants} but unreliable as independent annotators, and humans working alongside them are susceptible to automation bias.
\citet{DBLP:conf/icwsm/PangakisW25} observe that LLM annotators tend toward high recall but low precision, and that iterative human-in-the-loop prompt refinement only partially mitigates this imbalance.
Additionally, LLM annotation quality can degrade on non-English data \citep{pmlr-v239-mohta23a}, a concern relevant to our French extraction task.

Our framework sidesteps the confidence-based paradigm entirely, using \emph{agreement between independent models} rather than a single model's self-assessed uncertainty as the quality signal.
This design also mitigates anchoring bias, because the human jury is presented with two conflicting outputs rather than a single LLM suggestion to accept or reject.

\subsection{Consensus-Based Verification}
\label{subsec:consensus}

Recent work leverages cross-model consensus as a quality signal for LLM outputs.
\citet{zhang2025consensus} propose Consensus Entropy, a metric that quantifies agreement among multiple vision-language models and routes low-agreement OCR outputs to a stronger expert model---achieving fully automated quality control but offering no human fallback when all models share the same blind spot.
\citet{davoudi2025collective} show that inter-model majority voting can validate LLM answers without ground truth, confirming that cross-model agreement is a reliable quality signal, though their framework remains purely model-based.
\citet{tseng2025evaluating} introduce a multi-agent discussion protocol in which disagreeing LLMs share their reasoning and re-annotate over multiple rounds; this improves agreement but incurs substantial computational cost, and cases that exceed model capability receive no human support.
Most closely related to our work, \citet{yuan2025mchr} present MCHR, a framework where three LLMs independently annotate content and majority-vote disagreements are routed to human reviewers---the same consensus-then-escalate logic as our Layer~1, though our design uses strict pairwise agreement between two models rather than three-way majority voting, yielding a more conservative gate with a lower silent error rate at the cost of a higher escalation rate.

Our framework extends this line of work in three ways.
First, we employ a \emph{redundancy design}: two independent consensus-plus-human systems rather than one, enabling a second round of cross-system consensus that catches errors a single system would miss.
Second, the resulting second layer specifically targets \emph{human annotator error}---mistakes introduced by fatigue or subjectivity---which single-layer designs propagate to the final output unchecked.
Third, when consensus fails, we escalate to a \emph{domain expert} rather than a stronger model, which is essential for tasks where models exhibit systematic blind spots on domain-specific edge cases.
The framework requires no distributional priors or iterative calibration---only a simple agreement gate---making it directly applicable wherever two or more models can be run in parallel.

\section{Double Triangle Annotation Framework}
\label{sec:framework}

\subsection{Motivation and Overview}
\label{subsec:motivation}

Reliable evaluation of structured extraction systems on complex historical documents requires high-precision ground-truth data.
Traditional human annotation is expensive and difficult to scale, while fully automated pipelines suffer from hallucination.
We propose the \textbf{Double Triangle Annotation} framework, which combines the computational capacity of Multimodal Large Language Models (MLLMs) with targeted human oversight.

Rather than serving as primary annotators, humans act as \emph{conflict resolvers} within a statistically motivated filtering system.
The design rests on two key insights:
\begin{enumerate}
    \item \textbf{Low probability of coincident error.}
    When two architecturally independent models produce the same output for a given field, the probability that both are wrong in the same way is low. Model consensus therefore serves as a high-confidence proxy for correctness.
    \item \textbf{Mitigation of human fallibility.}
    A second layer compares two independent model--human systems against each other, catching errors introduced by annotator fatigue or subjectivity.
\end{enumerate}

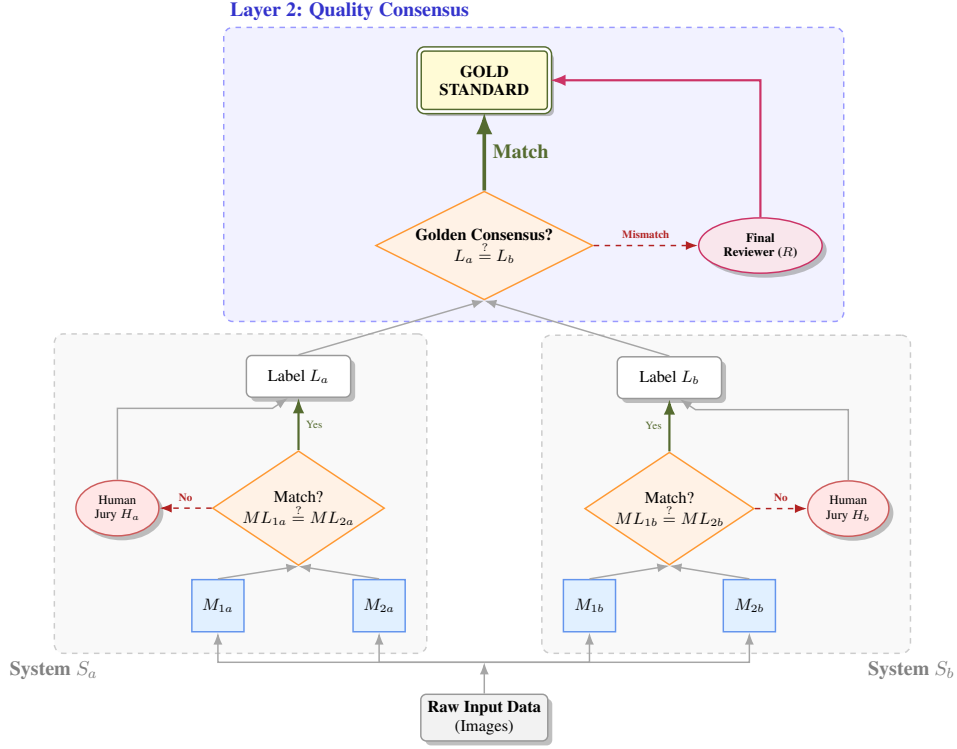
\begin{figure*}[t]
    \centering
    \resizebox{0.8\textwidth}{!}{
        \begin{tikzpicture}[
            node distance=1.0cm and 0.5cm,
            font=\sffamily,
            % --- STYLES ---
            data/.style={
                rectangle, rounded corners=3pt, draw=gray!80, fill=white,
                thick, minimum height=0.8cm, minimum width=2cm, align=center, drop shadow, font=\footnotesize
            },
            model/.style={
                rectangle, draw=effBorder, fill=effBlue, thick,
                minimum size=1cm, align=center, font=\footnotesize
            },
            human/.style={
                ellipse, draw=qcBorder, fill=qcRed, thick,
                minimum height=0.8cm, minimum width=1.5cm, align=center, drop shadow, font=\scriptsize
            },
            expert/.style={
                ellipse, draw=purple!80, fill=purple!10, thick, font=\bfseries\scriptsize,
                minimum height=1cm, minimum width=2cm, align=center, drop shadow
            },
            decision/.style={
                diamond, aspect=1.5, draw=orange!80, fill=orange!10, thick,
                align=center, font=\tiny, inner sep=1pt
            },
            golden/.style={
                rectangle, rounded corners=3pt,
                draw=goldenGreen, fill=yellow!20,
                thick, double, double distance=1.5pt,
                minimum height=1.2cm, minimum width=2.5cm,
                align=center, font=\bfseries\small, drop shadow
            },
            systemGroup/.style={
                draw=gray!40, dashed, fill=gray!5, rounded corners=5pt,
                inner sep=0.4cm, thick
            },
            layerGroup/.style={
                draw=blue!40, dashed, fill=blue!5, rounded corners=5pt,
                inner sep=0.4cm, thick
            },
            % Arrows
            flow/.style={->, >=LaTeX, thick, draw=gray!70},
            autoPath/.style={->, >=LaTeX, very thick, draw=goldenGreen},
            conflictPath/.style={->, >=LaTeX, thick, draw=conflictRed, dashed, font=\tiny\bfseries},
            expertPath/.style={->, >=LaTeX, very thick, draw=purple!80}
        ]

        % ==================== INPUT ====================
        \node[data, fill=gray!10] (raw) at (0,0) {\textbf{Raw Input Data}\\(Images)};

        % ==================== LAYER 1: SYSTEMS ====================

        % --- System A (Left) ---
        \node[decision, above left=3cm and 1.5cm of raw, minimum width=2.5cm, minimum height=1.5cm, font=\small] (dec1) {Match?\\$ML_{1a} \stackrel{?}{=} ML_{2a}$};
        \node[model, below left=0.8cm and 0.2cm of dec1] (m1a) {$M_{1a}$};
        \node[model, below right=0.8cm and 0.2cm of dec1] (m2a) {$M_{2a}$};
        \node[human, left=1cm of dec1] (jury1) {Human\\Jury $H_a$};
        \node[data, above=1cm of dec1] (out1) {Label $L_a$};

        % --- System B (Right) ---
        \node[decision, above right=3cm and 1.5cm of raw, font=\small] (dec2) {Match?\\$ML_{1b} \stackrel{?}{=} ML_{2b}$};
        \node[model, below left=0.8cm and 0.2cm of dec2] (m1b) {$M_{1b}$};
        \node[model, below right=0.8cm and 0.2cm of dec2] (m2b) {$M_{2b}$};
        \node[human, right=1cm of dec2] (jury2) {Human\\Jury $H_b$};
        \node[data, above=1cm of dec2] (out2) {Label $L_b$};

        % ==================== CONNECTIONS LAYER 1 ====================
        \draw[flow] (raw.north) -- ++(0,0.6) coordinate (split);
        \draw[flow] (split) -| (m1a.south);
        \draw[flow] (split) -| (m2a.south);
        \draw[flow] (split) -| (m1b.south);
        \draw[flow] (split) -| (m2b.south);

        \draw[flow] (m1a.north) -- (dec1.south); \draw[flow] (m2a.north) -- (dec1.south);
        \draw[flow] (m1b.north) -- (dec2.south); \draw[flow] (m2b.north) -- (dec2.south);

        \draw[autoPath] (dec1.north) -- node[right, font=\tiny, text=goldenGreen] {Yes} (out1.south);
        \draw[autoPath] (dec2.north) -- node[left, font=\tiny, text=goldenGreen] {Yes} (out2.south);

        \draw[conflictPath] (dec1.west) -- node[above, text=conflictRed] {No} (jury1.east);
        \draw[flow] (jury1.north) |- ($(out1.south)+(-0.5,-0.2)$) -- ($(out1.south)+(-0.2,0)$);

        \draw[conflictPath] (dec2.east) -- node[above, text=conflictRed] {No} (jury2.west);
        \draw[flow] (jury2.north) |- ($(out2.south)+(0.5,-0.2)$) -- ($(out2.south)+(0.2,0)$);

        % ==================== LAYER 2: QUALITY CONSENSUS ====================
        \coordinate (centerPoint) at (0, 5);
        \node[decision, aspect=2, font=\small, above=3cm of raw] (metaDec) at (0, 5) {\textbf{Golden Consensus?}\\$L_a \stackrel{?}{=} L_b$};

        \node[expert, right=2cm of metaDec] (expert) {Final\\Reviewer ($R$)};
        \node[golden, above=1.5cm of metaDec] (final) {GOLD\\STANDARD};

        \draw[flow] (out1.north) -- (metaDec.south);
        \draw[flow] (out2.north) -- (metaDec.south);

        \draw[autoPath, line width=2pt] (metaDec.north) -- node[right, font=\bfseries, text=goldenGreen] {Match} (final.south);
        \draw[conflictPath] (metaDec.east) -- node[above, text=conflictRed] {Mismatch} (expert.west);
        \draw[expertPath] (expert.north) |- (final.east);

        % ==================== GROUPING BOXES ====================
        \begin{scope}[on background layer]
            \node[systemGroup, fit=(out1) (jury1) (m1a) (m2a) (dec1)] (sysA) {};
            \node[below, font=\bfseries\color{gray}] at (sysA.south west) {System $S_a$};

            \node[systemGroup, fit=(out2) (jury2) (m1b) (m2b) (dec2)] (sysB) {};
            \node[below, font=\bfseries\color{gray}] at (sysB.south east) {System $S_b$};

            \coordinate (mirrorLeft) at ($(metaDec.west) + (-2.5,0)$);
            \node[layerGroup, fit=(metaDec) (final) (expert) (mirrorLeft)] (layer2) {};
            \node[above right, font=\bfseries\color{blue!60!gray}] at (layer2.north west) {Layer 2: Quality Consensus};
        \end{scope}
        \end{tikzpicture}
    }
    \caption{Overview of the Double Triangle Annotation framework. Layer~1 (gray boxes) pairs two independent models per system; consensus auto-accepts labels while conflicts are routed to a human jury. Layer~2 (blue box) cross-checks the two systems' outputs and escalates residual mismatches to a final reviewer.}
    \label{fig:double_triangle}
\end{figure*}

The objective is to produce a gold-standard dataset with near-perfect accuracy at a fraction of the cost of manual annotation.
Figure~\ref{fig:double_triangle} illustrates the full architecture.

% --- Double Triangle TikZ Figure (full-width) ---

\subsection{Layer 1: The Annotation Triangle}
\label{subsec:layer1}

The first layer maximizes throughput by leveraging cross-model consensus to automate labeling, reserving human labor strictly for ambiguous cases.

\subsubsection{Architectural Components}
Each annotation unit, termed a \textbf{System} ($S$), comprises three agents:
\begin{itemize}
    \item \textbf{Two machine annotators} ($M_1$, $M_2$): independent models (e.g., MLLMs or OCR engines) that extract structured text from document images in parallel.
    \item \textbf{One human jury} ($H$): an annotator who acts solely as a conflict resolver.
\end{itemize}

\subsubsection{Consensus-Based Workflow}
For every input image $I$, the process follows a verification-by-agreement logic:
\begin{enumerate}
    \item \textbf{Parallel inference.} $M_1$ and $M_2$ independently produce provisional labels $ML_1$ and $ML_2$.
    \item \textbf{Automated filtering.}
    \begin{itemize}
        \item \textbf{Consensus} ($ML_1 = ML_2$): the label is automatically accepted as the system output $L$, based on the low probability of coincident error (\S\ref{subsubsec:independence}).
        \item \textbf{Divergence} ($ML_1 \neq ML_2$): the case is flagged and routed to the human jury.
    \end{itemize}
    \item \textbf{Human adjudication.} The jury reviews the original image $I$ alongside both model predictions and determines the final label $L$, shifting the human role from \emph{producer} to \emph{verifier}.
\end{enumerate}

\subsubsection{The Independence Requirement}
\label{subsubsec:independence}
The consensus mechanism rests on a simple probabilistic argument: if two models fail independently, the chance that both produce the \emph{same} wrong output for the \emph{same} field is negligible.

Let $\epsilon_1$ and $\epsilon_2$ denote the per-field error rates of $M_1$ and $M_2$.
Under an independence assumption, the probability that both models err on a given field is $\epsilon_1 \cdot \epsilon_2$.
A consensus error---a \emph{silent} failure that passes the agreement gate---is even less likely, because it requires the two models to produce not merely wrong outputs, but the \emph{same} wrong output.
When two independently erring models produce \emph{different} incorrect strings, the gate catches the disagreement.
The silent error rate is therefore strictly bounded:
\begin{equation}
    P(\text{consensus error}) \leq \epsilon_1 \cdot \epsilon_2
    \label{eq:independence}
\end{equation}
We define the per-system silent error rate $\epsilon_S = P(\text{consensus error})$; from Equation~\ref{eq:independence}, $\epsilon_S \leq \epsilon_1 \cdot \epsilon_2$.
Because the human jury corrects all detected disagreements, $\epsilon_S$ is the dominant source of residual error within each Layer~1 system---and it is the quantity that Layer~2 further reduces through cross-system comparison (Equation~\ref{eq:cascaded_error}).

The bound in Equation~\ref{eq:independence} is conservative: the true rate is lower by a factor that depends on how many distinct wrong outputs a field admits.
With empirically observed error rates of $\epsilon_1 \approx 0.087$ and $\epsilon_2 \approx 0.044$ (Section~\ref{sec:experiments}), we obtain $\epsilon_S \leq 0.004$---fewer than 1 in 250 consensus-accepted fields is expected to contain an error, and the actual rate is likely lower still.

For this independence assumption to hold, $M_1$ and $M_2$ must be \textbf{architecturally independent} so that their failure modes do not correlate.
We identify two strategies:
\begin{itemize}
    \item \textbf{Cross-family}: pairing models from different providers (e.g., Llama vs.\ Claude) to ensure distinct training data distributions.
    \item \textbf{Cross-modal}: pairing a vision-based model (OCR) with a semantic model (MLLM), so that their error distributions are structurally disjoint.
\end{itemize}

\subsubsection{Verification Sampling}
To monitor the \emph{silent error rate}---errors that pass consensus undetected---we implement a stochastic quality check: a random subset of consensus-accepted cases is routed to the human jury to verify the ongoing validity of the independence assumption.

\subsection{Layer 2: The Validation Triangle}
\label{subsec:layer2}

To mitigate the subjectivity and fatigue risks inherent in a single human jury, the second layer operates as a meta-verification step.

\subsubsection{Meta-Architectural Components}
This layer treats each Layer~1 system ($S$) as a single annotator unit.
The structure comprises:
\begin{itemize}
    \item \textbf{Two independent systems} ($S_a$, $S_b$): distinct Layer~1 instances with different underlying model pairs, maintaining orthogonal independence at the system level.
    \item \textbf{One final reviewer} ($R$): a domain expert (e.g., a historian) acting as the ultimate arbiter.
\end{itemize}

\subsubsection{The Validation Workflow}
The process filters residual model and human errors through cross-system comparison:
\begin{enumerate}
    \item \textbf{Dual generation.} Systems $S_a$ and $S_b$ independently process the same data to produce refined labels $L_a$ and $L_b$.
    \item \textbf{Meta-comparison.}
    \begin{itemize}
        \item \textbf{Golden consensus} ($L_a = L_b$): the label is automatically designated as gold standard $G$, since two disjoint model pairs and potentially distinct human juries reached the same conclusion.
        \item \textbf{Conflict} ($L_a \neq L_b$): mismatches indicate complex edge cases or residual annotator error.
    \end{itemize}
    \item \textbf{Expert resolution.} Conflicting entries are routed to the final reviewer $R$, who determines the gold-standard label.
\end{enumerate}

\subsubsection{Error Probability Reduction}
By cross-checking decisions from independent systems with different human juries, the framework induces a cascaded failure model.
The probability of an error persisting in the final gold standard is bounded by the product of the two per-system error rates defined in Equation~\ref{eq:independence}:
\begin{equation}
    P(\text{error in } G) \leq \epsilon_{S_a} \times \epsilon_{S_b} \leq (\epsilon_{1a} \cdot \epsilon_{2a})(\epsilon_{1b} \cdot \epsilon_{2b})
    \label{eq:cascaded_error}
\end{equation}
Since each $\epsilon_S$ is already the product of two small error rates, their product approaches zero rapidly.
Expensive expert intervention is thereby allocated only to the small fraction of the dataset where the two systems disagree.

\section{Dataset: The Rosenwald Guides}
\label{sec:dataset}

\begin{figure}[t]
    \centering
    \includegraphics[width=\columnwidth]{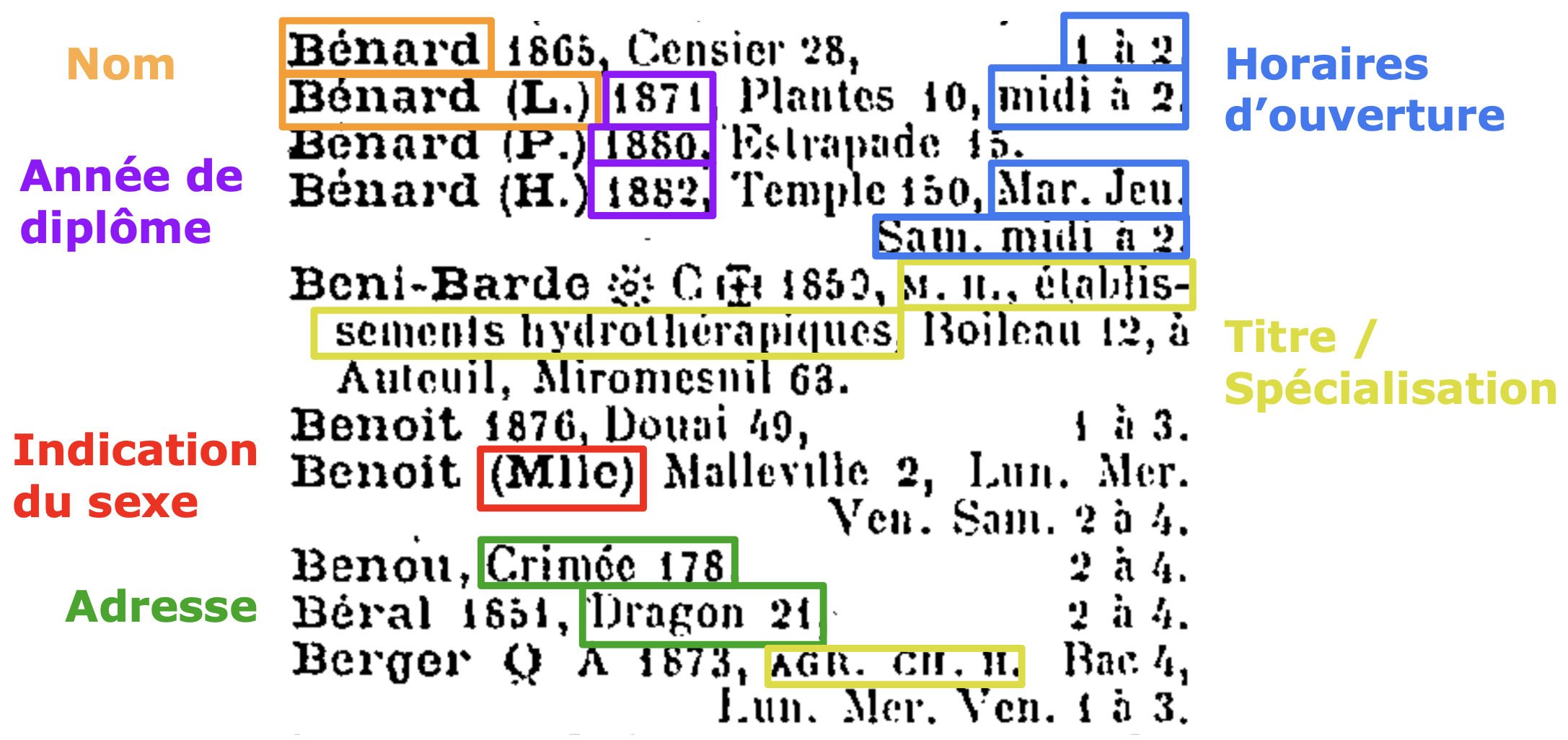}
    \caption{Annotated entries from the Rosenwald Guide (1887, p.~24). Each entry contains structured fields (name, year, notes, address, hours), but missing values and overlapping layouts introduce extraction noise.}
    \label{fig:close-look}
\end{figure}

The \emph{Guides Rosenwald} are French medical directories listing physicians and pharmacists across France and its colonies, published annually from 1887 to 1949 (47~volumes, each approximately one thousand pages).
As illustrated in Figure~\ref{fig:close-look}, each page lists doctor entries in a semi-structured format: a typical entry contains a name (\emph{nom}), diploma year (\emph{ann\'ee}), professional notes or specialization, address (\emph{adresse}), and consultation hours (\emph{horaires}).
Extraction is complicated by missing fields, overlapping entries, interspersed advertisements, and degrading print quality.

In this work, we process the sections titled \emph{Docteurs en m\'edecine et en chirurgie} and \emph{Officiers de sant\'e} for Paris and the d\'epartements, covering 20~volumes (1887--1906) with a total of 4{,}116~pages.
This subset provides sufficient scale and diversity---in typography, layout variation, and print degradation---to evaluate the Double Triangle annotation framework under realistic conditions.

\section{Experiments}
\label{sec:experiments}

We evaluate the Double Triangle framework along two axes: \emph{quality}---whether the two-layer design effectively corrects model and human errors---and \emph{efficiency}---whether consensus filtering successfully automates the majority of annotation work.

\subsection{Experimental Setup}
\label{subsec:setup}

\paragraph{Dataset.}
We conduct a controlled evaluation on page~32 of the 1887 Rosenwald Guide, which lists doctors of medicine and surgery in Paris.
The page contains 76~entries (40 in the left column, 36 in the right), yielding approximately 380 structured fields (name, address, consultation hours, etc.).
Two annotators independently produced manual transcriptions of this page without any model assistance; their reconciled output serves as the gold standard~$G$.

\paragraph{Model configuration.}
Following the independence requirement (\S\ref{subsubsec:independence}), we instantiate two systems with cross-provider model pairs:
\begin{itemize}
    \item \textbf{System $S_a$:} $M_{1a}$ = \texttt{claude-sonnet-4-5} \citep{anthropic2025sonnet45}, $M_{2a}$ = \texttt{qwen3-vl-235b} \citep{qwen2025qwen3vl}.
    \item \textbf{System $S_b$:} $M_{1b}$ = \texttt{llama4-maverick} \citep{meta2025llama4}, $M_{2b}$ = \texttt{grok-4-0709} \citep{xai2025grok4}.
\end{itemize}
We use MLLMs rather than traditional OCR engines, as pilot experiments showed OCR outputs were too noisy for reliable consensus comparison.

\paragraph{Human annotators.}
Jury $H_a$ is a research assistant fluent in French; jury $H_b$ is the project researcher with B1-level French proficiency.\footnote{B1 (``intermediate'') on the Common European Framework of Reference for Languages (CEFR).}
The final reviewer $R$ is the same person as $H_b$.
This represents a weaker configuration than the ideal three-independent-annotator design; the reported error rates therefore constitute an \emph{upper bound} on the errors expected under full independence, since a dedicated reviewer would be strictly more effective than $H_b$ serving dual roles.

\subsection{Evaluation Metrics}
\label{subsec:eval-metrics}

We evaluate all system outputs against the gold standard $G$.
For any hypothesis text $X$ (a model prediction, jury-corrected version, or reviewer-corrected version), we report four metrics.

\paragraph{Word Error Rate (WER).}
WER measures word-level edit distance:
\begin{equation}
    \mathrm{WER}(X, G) = \frac{S_w + D_w + I_w}{N_w},
    \label{eq:wer}
\end{equation}
where $S_w$, $D_w$, $I_w$ are word-level substitutions, deletions, and insertions, and $N_w$ is the word count of $G$.
We report WER \textbf{before} and \textbf{after} human correction at each layer.

\paragraph{Character Error Rate (CER).}
CER captures finer-grained errors:
\begin{equation}
    \mathrm{CER}(X, G) = \frac{S_c + D_c + I_c}{N_c},
    \label{eq:cer}
\end{equation}
where $S_c$, $D_c$, $I_c$ and $N_c$ are defined analogously at the character level.

\paragraph{Fields to correct.}
We quantify human workload at the field level.
Let $N_{\text{fields}}$ be the number of atomic fields in the dataset. For a given operator (jury or reviewer), we define:
\begin{equation}
    N_{\text{correct}} = \sum_{j=1}^{N_{\text{fields}}} \mathbb{I}\!\left[X_j \neq Y_j\right],
    \label{eq:fields-correct}
\end{equation}
where $X_j$ is the pre-review value for field $j$, $Y_j$ is the value after intervention, and $\mathbb{I}(\cdot)$ is the indicator function.

\paragraph{Human effort ratio.}
We normalize the correction count:
\begin{equation}
    \mathrm{EffortRatio} = \frac{N_{\text{correct}}}{N_{\text{fields}}} \times 100\%.
    \label{eq:effort-ratio}
\end{equation}
This metric captures the fraction of the dataset requiring manual edits at each layer.

\subsection{Results and Analysis}
\label{subsec:results}

Results are reported in Table~\ref{tab:res-triangle}.

% --- Results table ---
\begin{table}[t]
\centering
\small
\setlength{\tabcolsep}{3pt}
\renewcommand{\arraystretch}{1.15}
\resizebox{\columnwidth}{!}{
\begin{tabular}{llccccrr}
\toprule
\textbf{Human} & \textbf{Input} &
\multicolumn{2}{c}{\textbf{WER}} & \multicolumn{2}{c}{\textbf{CER}} &
\makecell{\textbf{Fields to}\\\textbf{correct}} &
\makecell{\textbf{Effort}\\\textbf{ratio}} \\
\cmidrule(lr){3-4}\cmidrule(lr){5-6}
& & Before & After & Before & After & & \\
\midrule
\multirow{2}{*}{$H_a$} & Claude & 0.087 & \multirow{2}{*}{0.030} & 0.053 & \multirow{2}{*}{0.018} & \multirow{2}{*}{70}  & \multirow{2}{*}{18.4\%} \\
                        & Qwen   & 0.044 &                        & 0.014 &                        &                       &                         \\
\midrule
\multirow{2}{*}{$H_b$} & Llama & 0.069 & \multirow{2}{*}{0.008} & 0.027 & \multirow{2}{*}{0.003} & \multirow{2}{*}{100} & \multirow{2}{*}{26.3\%} \\
                        & Grok  & 0.141 &                        & 0.050 &                        &                       &                         \\
\midrule
\multirow{2}{*}{$R$}   & $H_a$ & 0.030 & \multirow{2}{*}{\textbf{0.003}} & 0.018 & \multirow{2}{*}{\textbf{0.001}} & \multirow{2}{*}{16}  & \multirow{2}{*}{4.2\%}  \\
                        & $H_b$ & 0.008 &                        & 0.003 &                        &                       &                         \\
\bottomrule
\end{tabular}
}
\caption{Results of the Double Triangle framework on page~32 of the 1887 Rosenwald Guide (~380 fields). Layer~1 rows show jury corrections ($H_a$, $H_b$); the Layer~2 row shows the final reviewer ($R$). ``Before'' and ``After'' refer to error rates before and after human correction at the respective layer.}
\label{tab:res-triangle}
\end{table}

\subsubsection{Accuracy}
\label{subsubsec:accuracy}

\paragraph{First layer.}
After jury correction, WER decreases for both systems: from 0.087/0.044 to 0.030 for $S_a$, and from 0.069/0.141 to 0.008 for $S_b$.
CER decreases for $H_b$ but anomalously increases for $H_a$; inspection reveals that $H_a$ accidentally left all fields blank for one entry, contributing errors that the models had partially captured.
This incident underscores the value of the second-layer review for catching single-annotator lapses.

\paragraph{Second layer.}
The final reviewer $R$ further reduces both WER (to 0.003) and CER (to 0.001).
However, two errors persist in the final output, which we analyze below.

\paragraph{Layer-wise comparison.}
Table~\ref{tab:res-triangle} can be read as an implicit ablation across increasingly strong configurations.
A single model achieves WER between 0.044 and 0.141; a single annotation triangle (one model pair plus jury) reduces this to 0.008--0.030; the full double triangle achieves 0.003.
Each layer yields a substantial improvement: Layer~1 consensus filtering eliminates the majority of model errors, and Layer~2 cross-system verification catches residual human and model mistakes that any single system would miss.
This progressive reduction empirically validates the two-layer design and demonstrates that both layers contribute meaningfully to the final accuracy.

\subsubsection{Error Analysis}

Detailed inspection reveals two residual errors,
both traceable to genuine visual ambiguity in the
source images. Under such extreme ambiguity,
the source itself induces correlated errors:
when the printed character is objectively difficult
to resolve, any reader---human or machine---is
likely to arrive at the same incorrect
interpretation, effectively breaching the
independence assumption. These cases represent
an irreducible failure mode of any
consensus-based mechanism, since the quality
signal collapses precisely when all annotators
share the same degraded input.

\begin{figure}[t]
    \centering
    \begin{subfigure}[b]{0.48\columnwidth}
        \includegraphics[width=\linewidth]{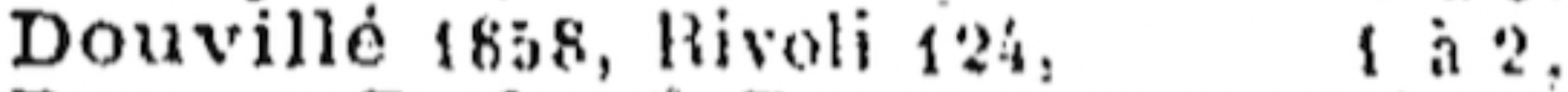}
        \caption{\emph{Douvill\'e}: accent resembles noise.}
    \end{subfigure}
    \hfill
    \begin{subfigure}[b]{0.48\columnwidth}
        \includegraphics[width=\linewidth]{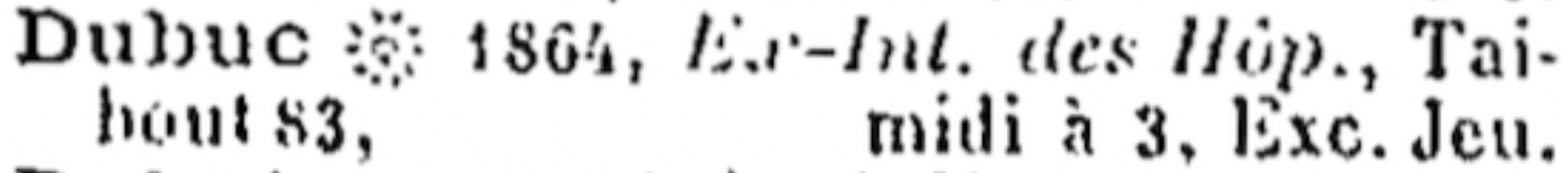}
        \caption{\emph{Taibout}: hyphen misread as \emph{t}.}
    \end{subfigure}
    \caption{The two residual errors both stem from genuine visual ambiguity that induces correlated errors across readers}
    \label{fig:error-cases}
\end{figure}

\begin{enumerate}
    \item \textbf{Character substitution.} The surname \emph{Douvill\'e} appears with an accent aigu that is printed as a faint dot, visually indistinguishable from noise (Figure~\ref{fig:error-cases}a).
    In $S_b$, Llama and Grok both recognize \emph{Douville} (missing accent), so $H_b$ has no disagreement signal and accepts the incorrect form.
    In $S_a$, Claude omits the entry entirely while Qwen also drops the accent; $H_a$ accidentally left all fields blank for this entry (the annotator lapse noted in \S\ref{subsubsec:accuracy}).
    The resulting discrepancy between $L_a$ and $L_b$ is escalated to $R$, who accepts the \emph{Douville} variant from $H_b$ without noticing the missing accent.

    \item \textbf{Character insertion.} The surname \emph{Taibout} contains a line-break hyphen after \emph{Tai-}, which models interpret as a faded character (Figure~\ref{fig:error-cases}b).
    In $S_a$, Claude and Qwen both produce \emph{Taitbout}, so the consensus gate auto-accepts the incorrect form without involving $H_a$.
    In $S_b$, Llama produces \emph{Taitbout} while Grok produces \emph{Taihout}---a different misreading that triggers the disagreement gate.
    However, $H_b$ resolves the conflict in favor of \emph{Taitbout}.
    Since $L_a = L_b$, the error passes Layer~2 undetected.
\end{enumerate}

% These cases illustrate three systematic failure modes:

% \paragraph{Correlated model errors.}
% Despite using cross-provider pairs, models can converge on the same mistake when the visual signal is genuinely ambiguous (e.g., faint diacritics, line-break artifacts).
% Agreement is therefore a necessary but not sufficient condition for correctness.

% \paragraph{Semantic correction bias.}
% LLMs may favor more frequent strings over faithful transcriptions. \emph{Douville} (a commune name) likely appears more often in training corpora than the surname \emph{Douvill\'e}; similarly, \emph{Rue Taitbout} (a Parisian street) is more common than the surname \emph{Taibout}.
% This frequency bias is particularly problematic for rare proper names and diacritics.

% \paragraph{Residual human subjectivity.}
% Even with two correction layers, annotators may default to the more plausible-looking candidate without verifying against the source image, especially under fatigue.
% Interface designs that prompt direct transcription rather than binary choice could mitigate this anchoring effect.

\subsubsection{Efficiency}

The consensus mechanism substantially reduces human workload.
In Layer~1, $H_a$ reviews 70 of 380 fields (18.4\%) and $H_b$ reviews 100 fields (26.3\%), meaning 74--82\% of fields are auto-accepted via model consensus.
In Layer~2, $R$ reviews only 16 fields (4.2\%).
The total human effort across both layers remains well below full manual annotation, and is distributed across three individuals ($H_a$, $H_b$, $R$), further reducing per-person workload and the risk of fatigue-induced errors.

\section{Benchmark Construction}
\label{sec:benchmark}

To demonstrate the scalability of the Double Triangle framework, we apply it to construct a benchmark from the Rosenwald Guides corpus described in \S\ref{sec:dataset}.

\subsection{Sampling and Preprocessing}

We apply stratified sampling over the 4{,}116~target pages, allocating pages per year proportional to each volume's share of the corpus.
This yields 105~sampled pages.
To improve model accuracy, we manually remove advertisements from each page and split the two-column layout into separate images, producing 210~column images.
The same cross-provider model pairs from \S\ref{subsec:setup} are used for annotation.

\subsection{Agreement and Filtering}

Table~\ref{tab:agreement} reports field-level agreement rates before and after quality filtering.
On the raw 210-column dataset, both model pairs achieve over 75\% field-level consensus, indicating that the majority of fields can be auto-accepted without human review.

A key advantage of the Double Triangle design is that Layer~1 consensus is computed \emph{before} any human correction, providing a cost-free signal of data quality.
This enables selective filtering: we retain only columns with fewer than 70~distinct fields, a field-level matching rate above 0.7, and a character-level matching rate above 0.6, discarding outlier columns where models performed poorly or the layout was unusually complex.
Because filtering occurs upstream of human effort, it further reduces annotation cost without compromising accuracy on the retained data.
The tradeoff is a potential selection bias toward easier columns, though the filtered set preserves the typographic and temporal diversity of the original stratified sample.

After filtering, 60~columns remain. Agreement rates rise to 87.6\% and 85.6\% for the two model pairs.

\begin{table}[t]
\centering
\small
\resizebox{\columnwidth}{!}{
\begin{tabular}{lrrrr}
\toprule
& \multicolumn{2}{c}{\textbf{Raw (210 cols)}} & \multicolumn{2}{c}{\textbf{Filtered (60 cols)}} \\
\cmidrule(lr){2-3}\cmidrule(lr){4-5}
\textbf{Model pair} & \textbf{Fields} & \textbf{\%} & \textbf{Fields} & \textbf{\%} \\
\midrule
Claude / Qwen & 37{,}314 / 46{,}830 & 79.7 & 11{,}294 / 12{,}895 & 87.6 \\
Llama / Grok  & 34{,}962 / 46{,}550 & 75.1 & 11{,}048 / 12{,}905 & 85.6 \\
\bottomrule
\end{tabular}
}
\caption{Field-level agreement rates by model pair, before and after quality filtering. ``Fields'' shows matched / total.}
\label{tab:agreement}
\end{table}

\subsection{Final Benchmark}

The resulting benchmark comprises 60~columns containing 13{,}595~structured fields.
In Layer~1, the consensus gate auto-accepts the majority of fields: jury $H_a$ reviews 1{,}601~divergent fields (12.4\%) and jury $H_b$ reviews 1{,}857 (14.4\%), while the remaining fields pass through without human involvement.
In Layer~2, only 991~fields (7.2\%) require correction by the final reviewer---the remaining 92.8\% are resolved by cross-system consensus alone.
Two annotators performed the corrections using a custom platform (Appendix~\ref{sec:appendix}) that highlights disagreement fields and marks differing characters, further reducing review time.
To monitor the silent error rate of consensus-accepted fields (\S\ref{subsubsec:independence}), we applied the verification sampling procedure: one randomly selected consensus-accepted field per column was routed to the human jury for independent verification.
Across the 60 sampled fields, no errors were detected.

This benchmark is, to our knowledge, the first structured-extraction ground truth for the Rosenwald Guides, and is made publicly available to support future work on historical document processing.

\section{Conclusion}
\label{sec:conclusion}

We presented the Double Triangle Annotation framework, a two-layer human-in-the-loop paradigm that uses cross-model consensus to automate the majority of annotation work while reserving human effort for genuinely ambiguous cases.
The framework rests on a single assumption---partial error independence between models---and requires no task-specific calibration or distributional priors.

On the Rosenwald Guides, the framework achieves a final WER of 0.003 and CER of 0.001, with model consensus auto-accepting over 85\% of the 13{,}595~fields in the at-scale benchmark.
The two residual errors both stem from genuine visual ambiguity in the source images, where the degraded input induces correlated errors that breach the independence assumption.
We release the resulting benchmark---the first structured-extraction ground truth for the Rosenwald Guides---to support future work on historical document processing.

Future directions include cross-modal model pairing (MLLM with OCR) to further reduce correlated failures, lightweight validation rules (e.g., address normalization) to catch systematic errors at the consensus stage, and progressive replacement of human juries with LLM-as-judge as model capability improves.

% REQUIRED for ACL: unnumbered, after Conclusion, before References.
% Does NOT count toward the 8-page content limit.
% Papers without this section will be desk-rejected.
\section*{Limitations}

\paragraph{Single-page controlled evaluation.}
The accuracy analysis in \S\ref{sec:experiments} is based on a single page (76~entries, ~380~fields). While the benchmark construction (\S\ref{sec:benchmark}) demonstrates scalability, the detailed error analysis and WER/CER measurements rely on a limited sample. Results may not generalize to pages with different layouts, denser entries, or greater print degradation.

\paragraph{Source ambiguity as an irreducible failure mode.}
In our error analysis (\S\ref{subsec:results}), the two residual errors reveal a distinct limitation: when the source image is genuinely ambiguous (e.g., faint diacritics, line-break artifacts), all readers---human and machine---converge on the same incorrect interpretation.
Such cases represent situations where extreme source
ambiguity induces correlated errors that breach the
independence assumption---an irreducible failure mode
that no consensus-based mechanism can detect, since the
degraded input constrains all readers toward the same
incorrect interpretation.

\paragraph{Annotator overlap.}
In our experimental setup, the final reviewer $R$ is the same person as jury $H_b$, weakening the independence between Layers~1 and~2. Although we argue that this configuration provides an upper bound on error rates (\S\ref{subsec:setup}), a fully independent three-annotator setup would provide stronger guarantees.

\paragraph{Benchmark filtering bias.}
The post-filtering step in benchmark construction (\S\ref{sec:benchmark}) removes columns with low model agreement, which may bias the benchmark toward easier cases. While filtering is a deliberate design choice that reduces human effort, the resulting benchmark may underrepresent the most challenging document layouts.

\paragraph{MLLM-only model pairs.}
All model pairs in our experiments use MLLMs. We do not evaluate the cross-modal configuration (MLLM paired with traditional OCR) discussed in \S\ref{subsubsec:independence}, which could provide structurally more independent error distributions at the cost of noisier baseline outputs.

\paragraph{No external framework comparison.}
We do not compare the Double Triangle framework against alternative annotation methodologies (e.g., another LLM-consensus based annotation framework, or standard human annotation).
Our evaluation relies on an inward comparison---single model vs.\ single triangle vs.\ double triangle---which validates the incremental value of each layer but does not directly benchmark against existing approaches.
A controlled comparison would require applying multiple annotation frameworks to the same dataset under comparable cost constraints, which we leave to future work.

% -----------------------------------------------------------------------
\section*{Acknowledgments}

J'aimerais remercier le professeur Jérôme Baudry de l'École Polytechnique Fédérale de Lausanne, qui m'a présenté le projet et m'a offert des conseils ainsi qu'un budget pour l'annotation tout au long du projet. C'est toujours un plaisir d'échanger avec vous. Merci à Mikhaël Moreau de l'Institut des humanités en médecine, qui a discuté en personne avec moi chaque semaine pour le projet « Quantifier l'invisible : Les femmes médecins dans les Guides Rosenwald ». Grâce à votre professionnalisme, je me suis senti respecté et encouragé pendant la collaboration. Merci à Amélie Puche de l'Institut des humanités en médecine, qui m'a guidé avant la rédaction du rapport. Je n'ai pas beaucoup d'expérience dans ce domaine, et votre aide est très importante pour moi. Merci également à Ella Bischoff de l'Université de Lausanne, qui a réalisé les annotations avec moi de manière très rigoureuse. Ce travail s'inscrit dans le cadre du projet FNS MEDIF, dirigé par la Prof. Aude Fauvel (IHM-UNIL/CHUV) et le Prof. Rémy Amouroux (IP-UNIL), projet FNS n° 215100 (\url{https://data.snf.ch/grants/grant/215100}).

% -----------------------------------------------------------------------
\bibliography{custom}

@inproceedings{NEURIPS2023_m5hisdoc,
  author = {Shi, Yongxin and Liu, Chongyu and Peng, Dezhi and Jian, Cheng and Huang, Jiarong and Jin, Lianwen},
  booktitle = {Advances in Neural Information Processing Systems},
  editor = {A. Oh and T. Naumann and A. Globerson and K. Saenko and M. Hardt and S. Levine},
  pages = {78483--78495},
  publisher = {Curran Associates, Inc.},
  title = {{M5HisDoc}: A Large-scale Multi-style {C}hinese Historical Document Analysis Benchmark},
  volume = {36},
  year = {2023}
}

@inproceedings{xu2019casia,
  author={Xu, Yue and Yin, Fei and Wang, Da-Han and Zhang, Xu-Yao and Zhang, Zhaoxiang and Liu, Cheng-Lin},
  booktitle={2019 International Conference on Document Analysis and Recognition (ICDAR)},
  title={{CASIA-AHCDB}: A Large-Scale {C}hinese Ancient Handwritten Characters Database},
  year={2019},
  pages={793--798},
  doi={10.1109/ICDAR.2019.00132}
}

@inproceedings{papadopoulos2013impact,
  author = {Papadopoulos, Christos and Pletschacher, Stefan and Clausner, Christian and Antonacopoulos, Apostolos},
  title = {The {IMPACT} dataset of historical document images},
  year = {2013},
  publisher = {Association for Computing Machinery},
  doi = {10.1145/2501115.2501130},
  booktitle = {Proceedings of the 2nd International Workshop on Historical Document Imaging and Processing},
  pages = {123--130},
  series = {HIP '13}
}

@inproceedings{clausner2015enp,
  author={Clausner, Christian and Pletschacher, Stefan and Antonacopoulos, Apostolos},
  booktitle={2015 13th International Conference on Document Analysis and Recognition (ICDAR)},
  title={The {ENP} image and ground truth dataset of historical newspapers},
  year={2015},
  pages={931--935},
  doi={10.1109/ICDAR.2015.7333898}
}

@inproceedings{simistira2016diva,
  author={Simistira, Foteini and Seuret, Mathias and Eichenberger, Nicole and Garz, Angelika and Liwicki, Marcus and Ingold, Rolf},
  booktitle={2016 15th International Conference on Frontiers in Handwriting Recognition (ICFHR)},
  title={{DIVA-HisDB}: A Precisely Annotated Large Dataset of Challenging Medieval Manuscripts},
  year={2016},
  pages={471--476},
  doi={10.1109/ICFHR.2016.0093}
}

@inproceedings{yan2010scut,
  author={Yan, Hanyu and Jin, Lianwen and Viard-Gaudin, Christian and Mouchere, Harold},
  booktitle={2010 12th International Conference on Frontiers in Handwriting Recognition},
  title={{SCUT-COUCH Textline\_NU}: An Unconstrained Online Handwritten {C}hinese Text Lines Dataset},
  year={2010},
  pages={581--586},
  doi={10.1109/ICFHR.2010.123}
}

@inproceedings{shen2020japanese,
  author = {Shen, Zejiang and Zhang, Kaixuan and Dell, Melissa},
  title = {A Large Dataset of Historical {J}apanese Documents with Complex Layouts},
  booktitle = {Proceedings of the IEEE/CVF Conference on Computer Vision and Pattern Recognition Workshops (CVPRW)},
  year = {2020}
}

@inproceedings{DBLP:conf/acl/DingQLCLJB23,
  author = {Bosheng Ding and Chengwei Qin and Linlin Liu and Yew Ken Chia and Boyang Li and Shafiq Joty and Lidong Bing},
  title = {Is {GPT-3} a Good Data Annotator?},
  booktitle = {Proceedings of the 61st Annual Meeting of the Association for Computational Linguistics (Volume 1: Long Papers)},
  pages = {11173--11195},
  publisher = {Association for Computational Linguistics},
  year = {2023},
  doi = {10.18653/v1/2023.acl-long.626}
}

@inproceedings{DBLP:conf/emnlp/WangLXZZ21,
  author = {Shuohang Wang and Yang Liu and Yichong Xu and Chenguang Zhu and Michael Zeng},
  title = {Want To Reduce Labeling Cost? {GPT-3} Can Help},
  booktitle = {Findings of the Association for Computational Linguistics: {EMNLP} 2021},
  pages = {4195--4205},
  publisher = {Association for Computational Linguistics},
  year = {2021},
  doi = {10.18653/v1/2021.findings-emnlp.354}
}

@inproceedings{DBLP:conf/eacl/KimMCRZ24,
  author = {Hannah Kim and Kushan Mitra and Rafael Li Chen and Sajjadur Rahman and Dan Zhang},
  title = {{MEGAnno+}: {A} Human-{LLM} Collaborative Annotation System},
  booktitle = {Proceedings of the 18th Conference of the European Chapter of the Association for Computational Linguistics: System Demonstrations},
  pages = {168--176},
  publisher = {Association for Computational Linguistics},
  year = {2024}
}

@inproceedings{DBLP:conf/chi/Wang0RMM24,
  author = {Xinru Wang and Hannah Kim and Sajjadur Rahman and Kushan Mitra and Zhengjie Miao},
  title = {Human-{LLM} Collaborative Annotation Through Effective Verification of {LLM} Labels},
  booktitle = {Proceedings of the {CHI} Conference on Human Factors in Computing Systems},
  pages = {303:1--303:21},
  publisher = {{ACM}},
  year = {2024},
  doi = {10.1145/3613904.3641960}
}

@inproceedings{DBLP:conf/icwsm/PangakisW25,
  author = {Nicholas Pangakis and Samuel Wolken},
  title = {Keeping Humans in the Loop: Human-Centered Automated Annotation with Generative {AI}},
  booktitle = {Proceedings of the Nineteenth International {AAAI} Conference on Web and Social Media},
  pages = {1471--1492},
  publisher = {{AAAI} Press},
  year = {2025},
  doi = {10.1609/icwsm.v19i1.35883}
}

@inproceedings{DBLP:conf/acl/SchroederRK25,
  author = {Hope Schroeder and Deb Roy and Jad Kabbara},
  title = {Just Put a Human in the Loop? {I}nvestigating {LLM}-Assisted Annotation for Subjective Tasks},
  booktitle = {Findings of the Association for Computational Linguistics: {ACL} 2025},
  pages = {25771--25795},
  publisher = {Association for Computational Linguistics},
  year = {2025}
}

@article{DBLP:journals/corr/abs-2503-06778,
  author = {Feng Gu and Zongxia Li and Carlos Rafael Colon and Benjamin Evans and Ishani Mondal and Jordan Lee Boyd-Graber},
  title = {Large Language Models Are Effective Human Annotation Assistants, But Not Good Independent Annotators},
  journal = {CoRR},
  volume = {abs/2503.06778},
  year = {2025},
  doi = {10.48550/arXiv.2503.06778},
  eprinttype = {arXiv},
  eprint = {2503.06778}
}

@inproceedings{pmlr-v239-mohta23a,
  author = {Jay Mohta and Kenan Ak and Yan Xu and Mingwei Shen},
  title = {Are Large Language Models Good Annotators?},
  booktitle = {Proceedings on ``I Can't Believe It's Not Better: Failure Modes in the Age of Foundation Models'' at NeurIPS 2023 Workshops},
  pages = {38--48},
  year = {2023},
  volume = {239},
  series = {Proceedings of Machine Learning Research},
  publisher = {PMLR}
}

@article{zhang2025consensus,
  author = {Yulong Zhang and Tianyi Liang and Xinyue Huang and Erfei Cui and Xu Guo and Pei Chu and Chenhui Li and Ru Zhang and Wenhai Wang and Gongshen Liu},
  title = {Consensus Entropy: Harnessing Multi-{VLM} Agreement for Self-Verifying and Self-Improving {OCR}},
  journal = {CoRR},
  volume = {abs/2504.11101},
  year = {2025},
  eprinttype = {arXiv},
  eprint = {2504.11101}
}

@article{yuan2025mchr,
  author = {Mingyue Yuan and Jieshan Chen and Zhenchang Xing and Gelareh Mohammadi and Aaron Quigley},
  title = {A Case Study of Scalable Content Annotation Using Multi-{LLM} Consensus and Human Review},
  journal = {CoRR},
  volume = {abs/2503.17620},
  year = {2025},
  eprinttype = {arXiv},
  eprint = {2503.17620}
}

@article{davoudi2025collective,
  author = {Seyed Pouyan Mousavi Davoudi and Amin Gholami Davodi and Alireza Amiri-Margavi and Alireza Shafiee Fard and Mahdi Jafari},
  title = {Collective Reasoning Among {LLMs}: A Framework for Answer Validation Without Ground Truth},
  journal = {CoRR},
  volume = {abs/2502.20758},
  year = {2025},
  eprinttype = {arXiv},
  eprint = {2502.20758}
}

@article{tseng2025evaluating,
  author = {Yu-Min Tseng and Wei-Lin Chen and Chung-Chi Chen and Hsin-Hsi Chen},
  title = {Evaluating Large Language Models as Expert Annotators},
  journal = {CoRR},
  volume = {abs/2508.07827},
  year = {2025},
  eprinttype = {arXiv},
  eprint = {2508.07827}
}

@techreport{anthropic2025sonnet45,
  author = {Anthropic},
  title = {Claude Sonnet 4.5 System Card},
  institution = {Anthropic},
  year = {2025},
  url = {https://www.anthropic.com/claude-sonnet-4-5-system-card}
}

@article{qwen2025qwen3vl,
  author = {{Qwen Team}},
  title = {Qwen3-{VL} Technical Report},
  journal = {CoRR},
  volume = {abs/2511.21631},
  year = {2025},
  eprinttype = {arXiv},
  eprint = {2511.21631}
}

@misc{meta2025llama4,
  author = {{Meta AI}},
  title = {The {Llama} 4 Herd: The Beginning of a New Era of Natively Multimodal {AI} Innovation},
  year = {2025},
  url = {https://ai.meta.com/blog/llama-4-multimodal-intelligence/}
}

@techreport{xai2025grok4,
  author = {{xAI}},
  title = {Grok 4 Model Card},
  institution = {{xAI}},
  year = {2025},
  url = {https://data.x.ai/2025-08-20-grok-4-model-card.pdf}
}

% -----------------------------------------------------------------------
% Appendix (after references, does not count toward page limit)
\appendix
\section{Annotation Platform}
\label{sec:appendix}

\begin{figure}[t]
    \centering
    \includegraphics[width=\columnwidth]{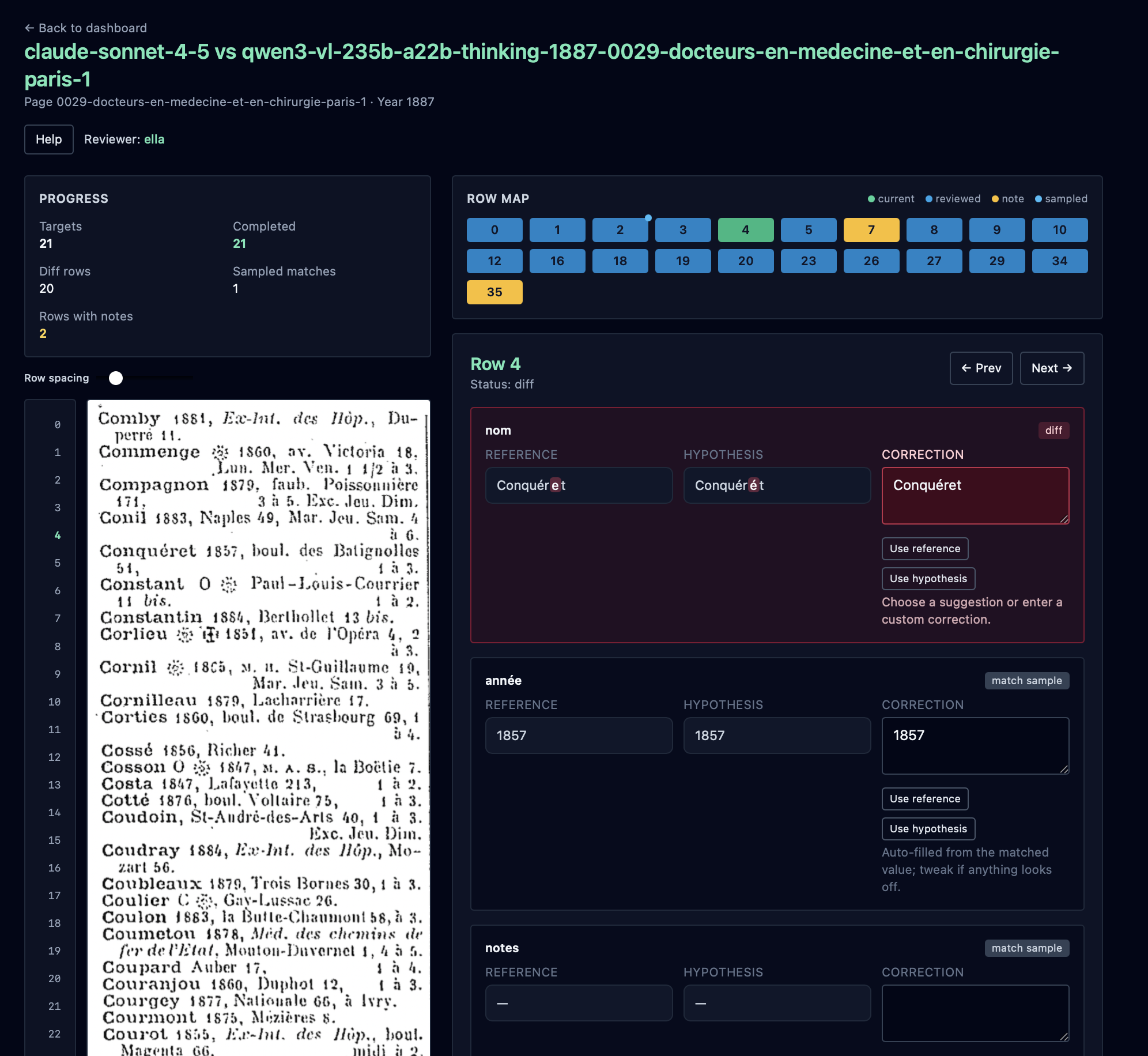}
    \caption{Screenshot of the annotation platform. The left panel shows the source document; the right panel presents model outputs side by side with a correction field. Character-level differences are highlighted to guide the reviewer.}
    \label{fig:platform}
\end{figure}

Figure~\ref{fig:platform} shows the annotation interface used by human juries and the final reviewer.
For each row flagged as a disagreement, the platform displays the original document image alongside both model outputs (``Reference'' and ``Hypothesis'') and a correction field.
Differing characters are highlighted to direct the annotator's attention to the specific point of conflict.

\section{Extraction Prompt}
\label{sec:prompt}

All four models (Claude, Qwen, Llama, Grok) receive the same prompt, reproduced below in its original French.
The prompt instructs the model to extract structured physician entries from a document image and return the results as tab-separated values.

\begin{lstlisting}
Vous analysez directement une image d'annuaire médical français du 19e et du 20e siècle.
L'entrée est une IMAGE d'UNE page complète d'annuaire médical historique.

IMPORTANT:
- L'image peut contenir des publicités, des textes d'éditeur, ou du texte non-médical en haut ou en bas.
- Ignorez-les complètement. Extrayez SEULEMENT les entrées de médecins.
- Lisez attentivement tous les textes visibles dans l'image, même s'ils sont partiellement effacés ou difficiles à lire.
- Si l'image ne contient AUCUNE entrée de médecin (seulement des publicités ou du texte non-médical), retournez une ligne vide dans le TSV.

Chaque entrée de médecin contient généralement :
- nom de famille (parfois avec préfixe ou parenthèses),
- année (diplôme ou année de référence),
- notes optionnelles (ex: "Ex-Int. des Hôp.", "AGR.", "M. A. M.", "CH. H.", "M. H."),
- adresse (rue, boulevard, avenue, etc.),
- horaires d'ouverture (abrégés: Lun., Mar., Mer., Jeu., Ven., Sam., Dim., "midi à 2", "2 à 4", etc.).

### TÂCHE
Analysez l'image et produisez un fichier TSV (Tab-Separated Values), où chaque ligne correspond à UNE entrée de médecin, avec ces colonnes séparées par des tabulations :

nom	année	notes	adresse	horaires

### DESCRIPTION DES COLONNES
- nom: Nom de famille du médecin (lisez l'orthographe exacte de l'image)
- année: Année de diplôme ou de référence (lisez exactement, ex: "1881" ou "1870")
- notes: Titres/affiliations hospitalières (ex: "Ex-Int. des Hôp.", "AGR.", "M. A. M.", "CH. H.")
- adresse: Adresse de rue complète (conservez accents/abréviations visibles)
- horaires: Informations horaires (ex: "Lun. Mer. Ven. 3 à 5")

### RÈGLES DE LECTURE D'IMAGE
- Lisez l'orthographe EXACTE visible dans l'image, accents, symboles (✶, AGR., etc.).
- Si un champ n'est pas visible ou illisible, laissez-le vide.
- N'inventez pas d'informations non visibles dans l'image.
- Ne manquez AUCUN médecin : parcourez systématiquement toute l'image pour extraire toutes les entrées présentes.
- ORDRE DE LECTURE CRITIQUE : L'image contient généralement DEUX COLONNES d'entrées de médecins. Respectez l'ordre de lecture suivant :
  1. Lisez la colonne de GAUCHE de haut en bas COMPLÈTEMENT
  2. Puis lisez la colonne de DROITE de haut en bas COMPLÈTEMENT
  3. NE MÉLANGEZ PAS les colonnes - terminez entièrement la gauche avant la droite
  4. Cet ordre est ESSENTIEL pour l'évaluation ultérieure
- Ne produisez que les entrées de médecins (ignorez publicités, textes d'éditeur).
- Incluez toujours les titres de civilité (Mme, Mlle, etc.) dans le champ nom s'ils sont visibles.
- Si aucune entrée de médecin n'est trouvée dans l'image, retournez seulement l'en-tête TSV.
- Séparez les colonnes par des TABULATIONS (	), pas des virgules.
- Aucun échappement nécessaire pour les virgules dans le texte.
- NE PAS entourer la sortie avec des balises markdown ou des blocs de code.
- Retournez UNIQUEMENT le contenu TSV brut, sans formatage markdown.

### CONSEILS POUR LA LECTURE D'IMAGE
- Examinez attentivement les zones de texte dense qui contiennent les listes de médecins.
- Les entrées de médecins sont généralement organisées en DEUX COLONNES distinctes sur la page.
- IMPORTANT : Suivez l'ordre de lecture colonne par colonne (gauche complète, puis droite complète).
- Faites attention aux abréviations typiques de l'époque (boul. pour boulevard, etc.).
- Les horaires peuvent être sur la même ligne ou sur la ligne suivante.
- Certains textes peuvent être en caractères plus petits ou partiellement effacés.
- Identifiez visuellement la séparation entre les deux colonnes pour éviter de mélanger l'ordre.

### EXEMPLE DE FORMAT ATTENDU:
nom	année	notes	adresse	horaires
Vallois	1848		St-André-des-Arts 50	2 à 4
Ravaux (Mme)	1883		Assomption 75	
Berline Hering (Mme)	1881	Halles 13	
Darcus Richardson	1861		Poisson 3	3 à 4, Exc. Dim
Benoit (Mlle)			Malleville 2	Lun. Mer. Ven. Sam. 2 à 4
Rigal 	1866	AGR. M. H.	Murillo 6	Lun. Mer. Ven. 1 à 3
Redard	1879	Méd. en chef des ch. de fer, Ex-Chef de Clin.	Constantinople 2	Lun. Mer. Ven. 5 à 6
Peltier (Henri)	1877		Marseille 2	4 à 6

### IMAGE À TRAITER
Analysez l'image ci-jointe et extrayez les données médicales selon les instructions ci-dessus.
\end{lstlisting}

\end{document}